\pdfoutput=1

\documentclass[11pt]{article}

\usepackage[]{naacl2021}

\usepackage{times}
\usepackage{latexsym}

\usepackage[T1]{fontenc}
\usepackage[utf8]{inputenc}

\usepackage{microtype}

\usepackage{graphicx}
\usepackage{multirow}
\usepackage{subcaption}
\usepackage{latexsym}
\usepackage{soul}
\usepackage{booktabs}
\usepackage{pifont}
\usepackage{xfrac}
\usepackage{arydshln}
\usepackage{amsfonts}
\usepackage{circledsteps}

\definecolor{myred}{HTML}{990000}
\definecolor{mygreen}{HTML}{009900}

\newcommand{\cmark}{\textcolor{mygreen}{\ding{51}}}%
\newcommand{\xmark}{\textcolor{myred}{\ding{55}}}%

\usepackage{xspace}
\usepackage{marvosym}
\makeatletter
\DeclareRobustCommand\onedot{\futurelet\@let@token\@onedot}
\def\@onedot{\ifx\@let@token.\else.\null\fi\xspace}

\def\eg{e.g\onedot} 
\def\ie{i.e\onedot}

 \def\vs{vs\onedot}

\def\resp{resp\onedot}
\makeatother

\definecolor{dimgray}{HTML}{696969}
\definecolor{gainsboro}{HTML}{dcdcdc}
\definecolor{lightergray}{HTML}{e6e6e6}

\definecolor{mylightblue}{HTML}{E0ECF7}
\definecolor{mydarkblue}{HTML}{092E6B}

\newcommand{\win}[1]{{\colorbox{mydarkblue}{\color{white}{\textbf{#1}}}}}
\newcommand{\lose}[1]{{\colorbox{mylightblue}{{\textbf{#1}}}}}

\newcommand{\abr}[1]{\textsc{#1}}
\newcommand{\accentor}{\abr{accentor}}
\newcommand{\dn}{{\accentor}}
\newcommand{\shane}[1]{{\color{blue}{\small\bf\sf [Shane: #1]}}}

\title{Adding Chit-Chat to Enhance Task-Oriented Dialogues}

\author{
    \textbf{Kai Sun} \textsuperscript{1}\thanks{\hspace{5pt}Work done as a research intern at Facebook. The code and data are available at \url{https://github.com/facebookresearch/accentor}.},
    \textbf{Seungwhan Moon}\textsuperscript{2},
    \textbf{Paul Crook}\textsuperscript{2},
    \textbf{Stephen Roller}\textsuperscript{3},
    \textbf{Becka Silvert}\textsuperscript{2}, \\
    \textbf{Bing Liu}\textsuperscript{2},    
    \textbf{Zhiguang Wang}\textsuperscript{2},
    \textbf{Honglei Liu}\textsuperscript{2},
    \textbf{Eunjoon Cho}\textsuperscript{2},
    and \textbf{Claire Cardie}\textsuperscript{1} \\
  \textsuperscript{1}Cornell University \\
\textsuperscript{2}Facebook,
  \textsuperscript{3}Facebook AI Research \\
  \Letter\hspace{3pt} {\tt ks985@cornell.edu}, {\tt shanemoon@fb.com} \\}

\begin{document}

\maketitle

\begin{abstract}

Existing dialogue corpora and models are typically designed under two disjoint motives: while task-oriented systems focus on achieving functional goals (\eg, booking hotels), open-domain chatbots aim at making socially engaging conversations. In this work, we propose to integrate both types of systems by Adding Chit-Chat to ENhance Task-ORiented dialogues ({\dn}), with the goal of making virtual assistant conversations more engaging and interactive. Specifically, we propose a Human $\leftrightarrow$ AI collaborative data collection approach for generating diverse chit-chat responses to augment task-oriented dialogues with minimal annotation effort. We then present our new chit-chat-based annotations to $23.8$K dialogues from two popular task-oriented datasets (Schema-Guided Dialogue and MultiWOZ 2.1) and demonstrate their advantage over the originals via human evaluation. Lastly, we propose three new models for adding chit-chat to task-oriented dialogues, explicitly trained to predict user goals \emph{and} to generate contextually relevant chit-chat responses. Automatic and human evaluations show that, compared with the state-of-the-art task-oriented baseline, our models can code-switch between task and chit-chat to be more engaging, interesting, knowledgeable, and humanlike, while maintaining competitive task performance.%
\end{abstract}

\section{Introduction}

With modeling innovations, increasing computing power, and a growing number of datasets, recent years have witnessed significant improvements in the performance of both task-oriented dialogue systems and chit-chat systems~\cite{adiwardana2020towards,roller2020recipes,hosseini2020simple,peng2020soloist}. Most research on dialogue systems focuses on a particular type of dialogue system. Work on task-oriented dialogue systems typically aims to track user goals with higher accuracy to better achieve functional goals~\cite{rastogi2019towards} with the sacrifice of not paying explicit attention to user experience, such as making the conversation more engaging, while the latter is usually the target of research on chit-chat systems~\cite{li2019acute}. In this work, we step forward and propose to integrate both types of systems by \textbf{A}dding \textbf{C}hit-\textbf{C}hat to \textbf{EN}hance \textbf{T}ask-\textbf{OR}iented dialogues ({\dn}), aiming to have a virtual assistant capable not only of performing various complex tasks such as checking the weather, booking hotels, and finding restaurants, but also incorporating casual and contextually relevant chit-chat. We hypothesize that the added chit-chat can make the assistant appear more social, personable, and engaging, without being misleading or inappropriate, compared with existing task-oriented dialogue systems.

\begin{figure}[t!]
   \begin{center}
   \includegraphics[width=0.46\textwidth]{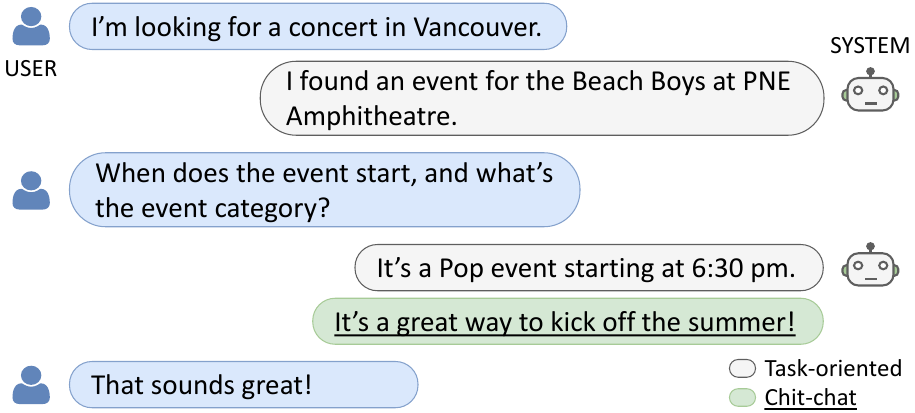}
   \end{center}
 \caption{A sample task-oriented dialogue snippet augmented by
 \textit{\ul{chit-chat}}.}
 \label{fig:sample}
\end{figure}

To show the feasibility of {\dn} and gather supervisory data for follow-up research, we propose a Human$\leftrightarrow$AI collaborative data construction approach that can effectively add suitable chit-chat to the beginning or end of system responses in existing task-oriented dialogue datasets. Specifically, we first generate chit-chat candidates for augmentation using off-the-shelf pre-trained language models and open-domain chatbots (Section~\ref{sec:candgen}). Next, we automatically filter out candidates that are unlikely to be of good quality using a filter model (Section~\ref{sec:candfilter}). Finally, human annotators label each of the remaining candidates as good or bad, with justifications (Section~\ref{sec:annotation}). We augment the Schema-Guided Dialogue (SGD)~\cite{rastogi2019towards} and MultiWOZ 2.1~\cite{eric2019multiwoz} corpora using the proposed approach. (See Figure~\ref{fig:sample} or Appendix~\ref{sec:dataset_examples} for examples.) We employ ACUTE-Eval~\cite{li2019acute} to compare the augmented versions with the originals along four axes: engagingness, interestingness, knowledge, and humanness. We find that the augmented dialogues are consistently preferred by human judges across the four axes for both datasets (Section~\ref{sec:dataeval}).

In addition, we propose and evaluate three models for adding chit-chat to task-oriented dialogues, including an end-to-end model and two code-switcher models built upon off-the-shelf task-oriented and chit-chat systems (Section~\ref{sec:approaches}). Compared with the baseline model trained with the original unaugmented data, our models trained with the augmented version can generate significantly higher-rated responses in terms of human preference while maintaining competitive task performance in goal tracking accuracy and action decision F1 (Section~\ref{sec:modeleval}).

\textbf{Our main contributions} are: we propose (1) a data augmentation approach for generating diverse chit-chat supervisory data for task-oriented dialogues, leveraging pre-trained generative models and a custom filter model to minimize human annotation effort; (2) new versions of the popular task-oriented datasets, SGD and MultiWOZ 2.1, with newly added chit-chat annotations to $23.8$K dialogues; and (3) three integrated chit-chat and task-oriented neural dialogue models for the above, substantially outperforming the state-of-the-art approach in terms of human evaluation of engagingness, interestingness, knowledge, and humanness.
To our knowledge, we are the first to propose an annotated dataset and models that study explicit code-switching between full-stack task-oriented dialogues and free-form chit-chat responses.

\section{Data Construction}
\label{sec:data}

\begin{figure}[t!]
   \begin{center}
   \includegraphics[width=0.48\textwidth]{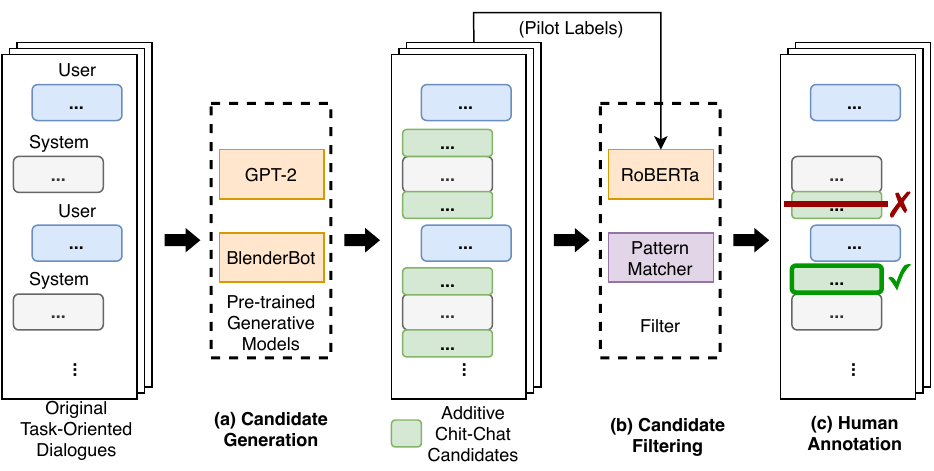}
   \end{center}
 \caption{Data construction overview: (a) We generate diverse free-form chit-chat candidates using the state-of-the-art pre-trained generative models to augment original task-oriented dialogues, and (b) filter out bad candidates using the custom filter to minimize annotation effort. (c) Crowd workers annotate contextually relevant chit-chat augmentation with justifications.}
 \label{fig:data_construction}
\end{figure}

In this section, we describe an approach to gather supervisory data for adding contextually relevant chit-chat to task-oriented dialogues. Our approach needs minimal annotation effort to augment suitable and diverse chit-chat add-ons that are not available in existing task-oriented datasets (Section~\ref{sec:relateddataset}). We primarily report results based on dialogues from the SGD dataset in this study, because it is the largest task-oriented dialogue dataset and is generally cleaner compared with most other task-oriented dialogue datasets. However, our approach is flexible and thus not limited to dialogues from a particular task-oriented dataset (Section~\ref{sec:dataeval}). Figure~\ref{fig:data_construction} shows the overview of our approach.

\subsection{Candidate Generation}
\label{sec:candgen}

Given a task-oriented dialogue $D=\{u_1,s_1,u_2,s_2,\ldots,u_n,s_n\}$, where $u_{1\ldots n}$ and $s_{1\ldots n}$ represent user turns and system turns, respectively, we generate chit-chat candidates for augmenting $s_i$ in two ways: (i) pass $u_1,s_1,\ldots,u_i,s_i$ to an off-the-shelf pre-trained model (a language model or a chit-chat chatbot) and let the model add tokens to the end of $s_i$; (ii) pass $u_1,s_1,\ldots,u_i$ to a pre-trained model and let the model generate a turn. We regard the output of (i) and (ii) as a chit-chat candidate to be appended and prepended to $s_i$, respectively. If a chit-chat candidate consists of multiple sentences, we also regard each individual sentence as a chit-chat candidate. We run differently sized GPT-2~\cite{radford2019language} and BlenderBot~\cite{roller2020recipes} with various decoding parameters as the pre-trained model and generate an average of $175.5$ candidates for each of the dialogues from the SGD dataset. See Appendix~\ref{sec:candsgendetails} for configuration details.

\begin{table*}[t!]
\centering
\scriptsize
\begin{tabular}{p{1.1cm}p{3.2cm}p{2.95cm}p{3.2cm}p{2.95cm}}
\toprule
 & \bf Appropriate Behaviours & \bf Examples & \bf Inappropriate Behaviours & \bf Examples \\
\midrule
\multirow{3}{1.1cm}{Opinions} & \multirow{3}{3.3cm}{Express general opinions about generic, impersonal, or non-sensitive topics.} & \multirow{3}{2.95cm}{- ``\textit{I love penguins.}''\newline - ``\textit{There's a lot of fun stuff to do.}''} & \multirow{3}{3.3cm}{Express strong personal opinions, or opinions on sensitive topics.} & \multirow{3}{2.95cm}{- ``\textit{I love you.}''\newline - ``\textit{The President is an idiot.}''} \\
\\\\
\cmidrule(lr){1-1} \cmidrule(lr){2-3} \cmidrule(lr){4-5}

\multirow{3}{1.1cm}{Preferences} & \multirow{3}{3.3cm}{Express preferences when making impersonal, or non-sensitive recommendations.} & \multirow{3}{2.95cm}{- ``\textit{Their latest album wasn't as good.}''\newline - ``\textit{Their food is good.}''} & \multirow{3}{3.3cm}{Express strong dispreferences, or preferences on personal or sensitive subjects.} & \multirow{3}{2.95cm}{- ``\textit{I hated it, but you might like it.}''\newline - ``\textit{Invite her! I like her better.}''} \\
\\\\
\cmidrule(lr){1-1} \cmidrule(lr){2-3} \cmidrule(lr){4-5}

\multirow{4}{1.1cm}{Physical Actions} & \multirow{4}{3.3cm}{Use epistemic verbs to express uncertainty or opinions, or refer through hearsay to actions that it may not perform.} & \multirow{4}{2.95cm}{- ``\textit{I hear it's beautiful.''}\newline - ``\textit{They say it tastes like chicken.}''} & \multirow{4}{3.3cm}{Behave as though it could act physically, or perform tasks outside of its role.}  & \multirow{4}{2.95cm}{- ``\textit{I haven't arrived there yet.}''\newline - ``\textit{I can drive you there.}''} \\
\\\\\\
\cmidrule(lr){1-1} \cmidrule(lr){2-3} \cmidrule(lr){4-5}

\multirow{4}{1.1cm}{Experiences} & \multirow{4}{3.3cm}{Refer to others' experiences or personify experiences it is capable of (\eg, reading).} & \multirow{4}{2.95cm}{- ``\textit{That sounds like a great trip!}''\newline - ``\textit{I enjoyed reading that novel.}''} & \multirow{4}{3.3cm}{Pretend to have experiences that it is incapable of.} & \multirow{4}{2.95cm}{- ``\textit{We didn't have that when I was a kid.}''\newline - ``\textit{My roommate used to eat there a lot.}''} \\
\\\\\\
\midrule
\multicolumn{5}{p{15cm}}{\textbf{Who is the virtual assistant?} This digital assistant is more than just a bot that spits out facts. It has access to a wide range of information which can express not only as factual commentaries but also as opinions and preferences. However, it is not a person and should not pretend to have real experiences or be capable of physical actions. It should be personable and personlike, without appearing counterfeit.} \\
\bottomrule
\end{tabular}
\caption{The role of the virtual assistant and its appropriate/inappropriate behaviors with examples.}
\label{tab:assistantdef}
\end{table*}

\subsection{Candidate Filtering}
\label{sec:candfilter}

We examine the quality of the model-generated candidates from Section~\ref{sec:candgen} by performing a pilot annotation ourselves on a small proportion of the candidates. The annotation results show that only about $\sfrac{1}{10}$ of the candidates are suitable. Therefore, instead of directly sending the candidates to crowd workers for annotation, we propose to build a filter model to automatically filter out candidates that are unlikely to be of good quality first to reduce potential annotation workload.

The filter is a hybrid model that consists of a RoBERTa-based binary classifier~\cite{liu2019roberta} and a rule-based ranker. The classifier takes as input an augmented dialogue, in which we explicitly surround the added chit-chat candidate with a pair of special tokens to help the model locate the candidate. We train the classifier with $1.7$K candidates that are labeled as good/bad from the pilot annotation. The rule-based ranker ranks each candidate based on (i) the posterior probability output by the binary classifier, (ii) whether the candidate matches a list of bad patterns (\eg, containing an URL), (iii) the frequency of appearances of the candidate among all generated candidates, (iv) the similarity to the other candidates for the dialogue, and (v) the similarity to the system response being augmented. While (i) and (ii) directly help evaluate the quality of the candidate, (iii), (iv), and (v) additionally help create more variety (\eg, punishing high-frequency candidates such as ``\textit{You're welcome}''). We keep the top ten candidates for each of the dialogues. We present more details in Appendix~\ref{sec:filterdetails}.

\subsection{Annotation}
\label{sec:annotation}

We ask annotators (crowd workers) to label each of the remaining candidates from Section~\ref{sec:candfilter} as \emph{good} or \emph{bad}. Additionally, to guide the annotation process, improve the potential quality, and facilitate the candidate distribution analysis, we also ask annotators to choose from four justifications that we come up with based on our pilot annotation experience to support their annotations. Annotators can choose one, both, or neither of the following justifications for a \emph{bad} candidate:
\begin{itemize}
    \item Inappropriate: The candidate does not fit into the context (\eg, repeating, unnatural), or it contradicts the context or the role of the assistant (Table~\ref{tab:assistantdef}). This category comprises most of the commonly found bad cases such as improper switching, providing opinions or comments that are incompatible with the context, and misusing verbal routine. 
    \item Misleading: The candidate provides additional information that is false or cannot be verified immediately. For example, the underlined candidate in the two-turn dialogue ``\textit{\textbf{U}: I want to book a hotel room in San Diego with a check in on Thursday. \textbf{A}: There are over 10 hotels in San Diego. \ul{I would stay at Arlo NoMad if I were you.}}'' should be marked as \emph{misleading} because ``\textit{Arlo NoMad}'' is newly introduced information, which the annotator would have to look up to verify that a hotel by this name exists in San Diego, even though the information may be true.
\end{itemize}
Annotators can choose one, both, or neither of the following justifications for a \emph{good} candidate:
\begin{itemize}
    \item Social: The candidate keeps the conversation flowing smoothly by appropriately switching to relevant topics, asking casual follow up questions, or engaging in social pleasantries. The design of this subcategory is inspired by the line of research that studies different social and discourse strategies in chit-chat dialogue systems~\cite{yu2016strategy}.
    \item Useful: The candidate enhances the conversation by appropriately offering opinions, commentaries, or pertinent and truthful information. Truthfulness should be established by conversational context or real world knowledge. To reduce annotation workload, if annotators have to use external resources (\eg, Wikipedia, search engines, maps) to verify information, they are instructed to label the candidate as \emph{misleading} instead. The design of this subcategory is inspired by the line of work on knowledge-grounded dialogue systems that study contextual knowledge injections~\cite{dinan2018wizard}.
\end{itemize}
We instruct annotators to evaluate each candidate independently as if it were the only augmentation for its associated dialogue. We discuss the additional dimension of complexity introduced by having multiple augmentations jointly in Section~\ref{sec:dataeval}.

\begin{table}[ht!]
\centering
\scriptsize
\begin{tabular}{lr}
\toprule
 \bf Metric & \bf Value \\ 
 \midrule
 \# of candidates & $228,250$ \\
 \# of unique candidates & $68,406$ \\
 vocabulary size & $10,005$ \\
 \# of distinct 2-grams & $59,259$ \\
 \# of distinct 3-grams & $131,989$ \\
 \# of distinct 4-grams & $195,508$ \\
 \# of distinct 5-grams & $239,278$ \\
 average length (in tokens) & $8.7$\\
 \midrule
  \# of good candidates (\%) & $94,600$ ($41.4$) \\
 ~$\diamond$ social & $86,324$ ($37.8$) \\
 ~$\diamond$ useful & $7,681$ (\phantom{0}$3.4$) \\
 ~$\diamond$ social \& useful & $577$ (\phantom{0}$0.3$) \\
 ~$\diamond$ other (good) & $18$ (\phantom{0}$0.0$) \\
 \midrule
  \# of bad candidates (\%) & $133,650$ ($58.6$) \\
 ~$\diamond$ inappropriate & $127,648$ ($55.9$) \\
 ~$\diamond$ misleading & $5,800$ (\phantom{0}$2.5$) \\
 ~$\diamond$ inappropriate \& misleading & $164$ (\phantom{0}$0.1$) \\
 ~$\diamond$ other (bad) & $38$ (\phantom{0}$0.0$) \\
\bottomrule
\end{tabular}
\caption{Statistics of annotated chit-chat candidates in {\dn}-SGD.}
\label{tab:statjustification}
\end{table}

Annotation time per dialogue is $243$s. The Fleiss' Kappa among crowd workers is $0.52$. We view the agreement score as reasonable since whether an added chit-chat candidate leads to improved quality of a conversation can be highly subjective in many scenarios. We denote our augmented version of the SGD dataset as {\dn}-SGD and summarize the statistics in Table~\ref{tab:statjustification}. We observe that the four provided justification categories provide adequate coverage of the justifications for most annotations. $41.4\%$ of the candidates are good, showing the effectiveness of candidate filtering. An analysis based on linguistic features suggests that bad candidates are more personal and negative than good candidates. Specifically, $40.0\%$ of bad candidates involve first-person pronouns, while the ratio is $26.5\%$ for good candidates. $81.7\%$ of good candidates have positive sentiment, measured by VADER, a lexicon and rule-based sentiment analysis tool~\cite{hutto2014vader}, while the ratio is $73.0\%$ for bad candidates.
Examples of the resulting dataset are presented in Appendix~\ref{sec:dataset_examples}.

\section{Approaches}
\label{sec:approaches}

\subsection{Task Formulations}

Since oracle information (\ie, oracle belief states and oracle action decisions) is not available in practical use and the SGD dataset does not have the associated database (\ie, a table of possible entities) released, we focus on exploring the end-to-end setting in which we generate \emph{delexicalized} task-oriented responses without using oracle information and database search results following \citet{hosseini2020simple}.
Given dialogue history (\ie, previous turns) as context, the goal of the model for each system turn is to accurately generate belief states (\ie, a list of $(domain, slot, value)$ triplets), action decisions (\ie, a list of $(domain, action\_type, slot)$ triplets), and a corresponding system response that is functionally accurate \emph{and} socially engaging. %

\subsection{Models}
\label{sec:models}

\begin{figure}[t!]
   \begin{center}
   \includegraphics[width=0.46\textwidth]{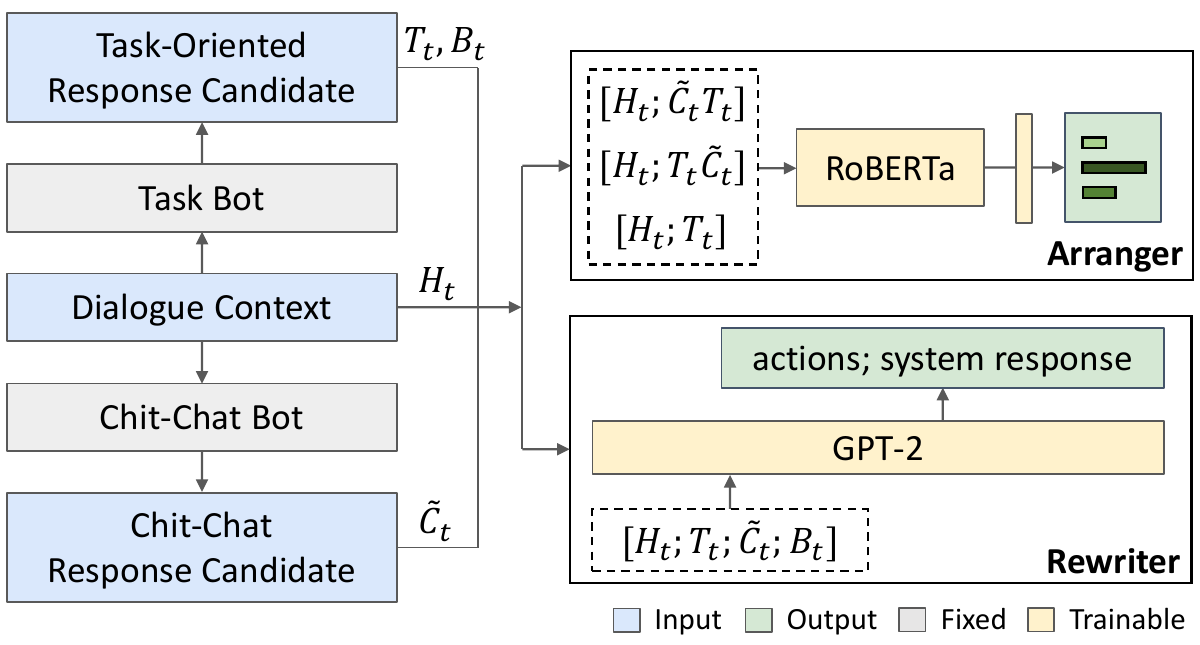}
   \end{center}
 \caption{A diagram for the proposed code-switching models. Given the dialogue context ($H_t$) and the pre-generated task-oriented and chit-chat response candidates  ($T_t, \tilde{C}_t$), the \textbf{Arranger} learns the optimal code-switching sequences (discriminative), while the \textbf{Rewriter} outputs free-form paraphrases (generative).}
 \label{fig:codeswitchers}
\end{figure}

We re-implement SimpleTOD~\cite{hosseini2020simple} as our main baseline model, which is a state-of-the-art model in the end-to-end setting we explore.
In addition, we propose an extension of SimpleTOD that incorporates chit-chat acts, as well as two new models (Arranger and Rewriter; Figure~\ref{fig:codeswitchers}) that code-switch between chit-chat and task-oriented responses more explicitly.

\paragraph{SimpleTOD.} It is a causal language model that models the joint probability over the concatenation of dialogue history $H_t$, belief states $B_t$, action decisions $A_t$, and a task-oriented response $T_t$ for each turn $t$. During inference, the model takes as input $H_t$ and generates $B_t$, $A_t$, and $T_t$. We refer readers to \citet{hosseini2020simple} for more details.

\paragraph{SimpleTOD+.} We extend SimpleTOD by introducing to the construction of input sequences a special new dialogue action \texttt{chit-chat} and good chit-chat candidates during training. Specifically, let $\mathbb{C}^+_t$ denote the set of good candidates for system turn $t$. If $\mathbb{C}^+_t$ is empty, we construct the same training sequence as SimpleTOD. Otherwise, for each $C_t \in \mathbb{C}^+_t$ that is labeled as a candidate to be prepended (\resp appended) to the turn, we use the concatenation of $H_t$, $B_t$, [\texttt{chit-chat}], $A_t$, $C_t$, and $T_t$ (\resp $H_t$, $B_t$, $A_t$, [\texttt{chit-chat}], $T_t$, and $C_t$) as a training sequence. 

\paragraph{Arranger.} This model arranges the output of an off-the-shelf task-oriented dialogue model and an off-the-shelf chit-chat model without intervening in the task. It outputs the belief states and action decisions generated by the task-oriented model without modification. To generate a response for each system turn $t$, this model takes as input (i) dialogue history $H_t$, (ii) a chit-chat response $\tilde{C}_t$ generated by the chit-chat model based on $H_t$, and (iii) a task-oriented response $T_t$ generated by the task-oriented dialogue model based on $H_t$. The model chooses one of the following as the response: $\tilde{C}_t$ followed by $T_t$, $T_t$ followed by $\tilde{C}_t$, and $T_t$ only. Specifically, the model encodes the concatenation of $H_t$ and each of these three responses by a RoBERTa encoder~\cite{liu2019roberta} and passes the resulting representations through a linear plus softmax layer to make the choice. To train the model, we form training instances by regarding each chit-chat candidate for turn $t$ from the training set of {\dn}-SGD as $\tilde{C}_t$ and the ground-truth task-oriented response as $T_t$ and setting the target choice based on the label (\ie, good/bad) and position (\ie, beginning/end of the response) of the candidate. 

\paragraph{Rewriter.} This model rewrites the output of an off-the-shelf task-oriented dialogue model and an off-the-shelf chit-chat model. It directly outputs the task-oriented model's belief states without modification and generates action decisions and a system response by a causal language model. The causal language model differs from SimpleTOD+ in that it has two additional components $T_t$ and $\tilde{C}_t$ added between $H_t$ and $B_t$ in each training sequence, where we form $T_t$ and $\tilde{C}_t$ in the same way as we do for Arranger. During the inference stage, it takes as input $H_t$, $T_t$ output by the task-oriented dialogue model, $\tilde{C}_t$ output by the chit-chat model, and $B_t$ output by the task-oriented dialogue model, and generates action decisions and a system response. Note that since $25.4\%$ of the annotated system turns in the training set of {\dn}-SGD have both good and bad chit-chat candidates, $\mathbb{C}_t^+$ can be non-empty when $\tilde{C}_t$ is a bad candidate, which enables the model to potentially generate a suitable chit-chat augmented response even if the output of the off-the-shelf chit-chat model is not good.

\subsection{Implementation Details} Unless specified otherwise, for causal language models, we use the $12$-layer GPT-2 ($117$M parameters) as the pre-trained language model~\cite{radford2019language} and fine-tune for ten epochs. We set the batch size to $36$ and the learning rate to $1\times 10^{-3}$. We employ the SimpleTOD baseline as the off-the-shelf task-oriented dialogue model for Arranger and Rewriter. We fine-tune a 90M parameter model~\cite{shuster2020dialogue} on each of the good chit-chat candidates with the associated dialogue history as the context from the training set of {\dn}-SGD following hyperparameters employed by \citet{roller2020recipes} and employ the resulting model as the off-the-shelf chit-chat model in Arranger and Rewriter. We use RoBERTa\textsubscript{BASE}~\cite{liu2019roberta} as the pre-trained language model for Arranger and fine-tune for three epochs with a learning rate of $2\times 10^{-5}$ and a batch size of $24$. %

\section{Experiments and Discussions}

\subsection{Data Evaluations}
\label{sec:dataeval}

\begin{table*}[ht!]
\centering
\scriptsize
\begin{tabular}{lcccccccccc}
\toprule
  
  & \multicolumn{2}{c}{\bf Joint GA} & \multicolumn{2}{c}{\bf Avg GA} & \multicolumn{2}{c}{\bf Act-Slot F1} & \multicolumn{2}{c}{\bf BLEU-4\textsubscript{orig}} & \multicolumn{2}{c}{\bf BLEU-4\textsubscript{aug}} \\
  & All & Seen & All & Seen & All & Seen & All & Seen & All & Seen \\
 \midrule
SimpleTOD & $29.4$ & $79.0$ & $46.9$ & $90.3$ & $61.7$ & $88.9$ & $12.3$ & $17.0$ & $8.0$ & $9.9$  \\
SimpleTOD+ & $29.3$ & $77.4$ & $47.5$ & $90.0$ & $61.5$ & $88.1$ & $11.2$ & $15.3$ & $10.8$ & $12.8$  \\
Arranger & $29.4$ & $79.0$ & $46.9$ & $90.3$ & $61.7$ & $88.9$ & $9.6$ & $13.5$ & $12.2$ & $14.4$  \\
Rewriter & $29.4$ & $79.0$ & $46.9$ & $90.3$ & $60.8$ & $86.6$ & $10.4$ & $14.5$ & $11.3$ & $13.3$  \\
\bottomrule
\end{tabular}
\caption{Automatic evaluation results on the test set of {\dn}-SGD.} %
\label{tab:autoeval}
\end{table*}

\paragraph{{\dn}-SGD.} We first evaluate {\dn} at the dataset level, aiming to answer two questions: Q1. Are task-oriented dialogues augmented with good chit-chat more preferred by human judges than the unaugmented? Q2. Does the answer to Q1 depend on how frequently we augment system responses with chit-chat? To answer these questions, we randomly sample $100$ dialogues from {\dn}-SGD, each having at least $8$ turns and enough candidates labeled as good for augmenting over $40\%$ of system responses so that we can compare the same task-oriented dialogue with different chit-chat injection frequencies that fall into each of the following four intervals: $(0.1, 0.2]$, $(0.2, 0.3]$, $(0.3, 0.4]$, and $(0.4, 1]$. Particularly, for the last interval, we augment all system responses that have chit-chat candidates labeled as good, while for the first three intervals, we only augment a randomly selected fraction to fit the interval. We employ ACUTE-Eval~\cite{li2019acute} for evaluation, whereby we ask human evaluators to make pairwise comparisons of complete dialogues over four axes: engagingness, interestingness, knowledge, and humanness. We provide the wording of the questions in Appendix~\ref{sec:acuteevalqs}.

\begin{figure}[h!]
   \begin{center}
   \includegraphics[width=0.45\textwidth]{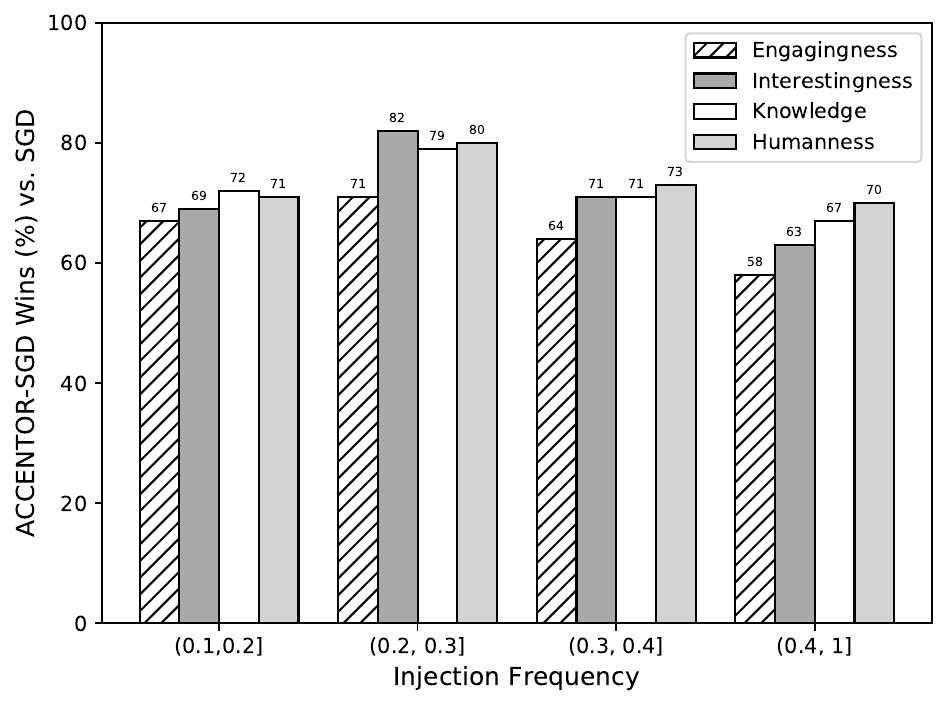}
   \end{center}
 \caption{Comparisons between SGD and {\dn}-SGD with different injection frequencies at the dataset level using ACUTE-Eval.}
 \label{fig:injfreq}
\end{figure}

As shown in Figure~\ref{fig:injfreq}, the chit-chat augmented dialogues from {\dn}-SGD are more preferred by human judges than the originals over all ACUTE-Eval metrics, regardless of the injection frequency ($\text{all p-values}<0.05$). Among different injection frequency ranges, $(0.2, 0.3]$ is the best. We offer three hypotheses to explain this finding: (i) $(0.2, 0.3]$ best balances being engaging and not too talkative. (ii) There are inevitable annotation errors, and scenarios where whether a candidate is good or bad is subjective. A higher injection frequency means a higher chance of being affected by these factors. (iii) Since candidates are labeled independently, inter-candidate incompatibility may arise (\eg, expressing contradicted preferences), especially when we have a high injection frequency. Table~\ref{tab:supinjfreq} shows a real example to support our hypotheses. Specifically, \Circled{3} is labeled as good but is indeed not a suitable (or at least a questionable) candidate, supporting the hypothesis (ii). While \Circled{2} and \Circled{4} are good candidates when we evaluate them separately, they may be less preferred if we assess them jointly because they convey the same meaning: ``\textit{Long Shot is a good comedy}.'' Having them together may appear incompatible (\ie, repetition) or sound verbose to the user, supporting the hypothesis (i) and (iii). 

\begin{table}[t!]
\centering
\scriptsize
\begin{tabular}{p{0.2cm}p{6.5cm}}
\toprule
\textbf{U}: & I like to find some movies directed by Jonathan Levine. \\
\textbf{A}: & What type of movies you interested? \\
\textbf{U}: & I like comic movies which is great. \\
\textbf{A}: & {\Circled{1}}~\textit{\ul{That's cool.}} There are 1 movies of your interest. What is your opinion about Long Shot? \\
\textbf{U}: & Long Shot is a good one. \\
\textbf{A}: & {\Circled{2}}~\textit{\ul{I agree.}} Do you like to run the movie? \\
\textbf{U}: & I need to run the movie with subtitles. \\
\textbf{A}: & Please give confirmation details: Running Long Shot with subtitles. {\Circled{3}}~\textit{\ul{Thank you for your help.}} \\
\textbf{U}: & Yes. Okay fine. \\
\textbf{A}: & The movie started playing. {\Circled{4}}~\textit{\ul{It's a great comedy.}} \\
\textbf{U}: & I want to make a reservation at a restaurant around there. \\
\textbf{A}: & Do you have any choice of restaurant and city and what is a booking time you need? \\
\textbf{U}: & I need it in San Francisco at 17:30 and the restaurant is Aliment. \\
\textbf{A}: & Please give confirmation details: Table booking at Aliment in San Francisco at 5:30 pm on march 1st and booking is for 2 people. {\Circled{5}}~\textit{\ul{Thank you.}} \\
\textbf{U}: & Fine. This is okay for me. \\
\textbf{A}: & Your booking is successful and location address is 786 Bush street. {\Circled{6}}~\textit{\ul{I hope you will enjoy the rest of your evening.}} \\
\textbf{U}: & Thank you so much for your help. \\
\textbf{A}: & {\Circled{7}}~\textit{\ul{You are very welcome.}} Do you need any other help? \\
\textbf{U}: & No, thanks a lot. \\
\textbf{A}: & Have a great day. \\
\bottomrule
\end{tabular}
\caption{An augmented dialogue (with injection frequency in $(0.4,1]$) that is less preferred than the unaugmented in terms of human evaluation (U: user; A: assistant; \textit{\ul{chit-chat}} is marked by circled numbers).} 
\label{tab:supinjfreq}
\end{table}

\begin{figure*}[ht!]
\centering

\begin{subfigure}[b]{.46\linewidth}
\setlength{\tabcolsep}{3pt}
\centering
\scriptsize
\begin{tabular}{rr|cccc}
& & \multicolumn{4}{c}{Win \%}\\
& &  SimpleTOD & SimpleTOD+ & Arranger & Rewriter \\
\midrule
\parbox[t]{2mm}{\multirow{4}{*}{\rotatebox[origin=c]{90}{Loss \%}}}
& SimpleTOD & ~~--\phantom{$^{\ast\ast}$} & ~~\win{63}$^{\ast\ast}$ & ~~\win{76}$^{\ast\ast}$ & ~~\win{91}$^{\ast\ast}$ \\
 & SimpleTOD+ & ~~\lose{37}$^{\ast\ast}$ & ~~--\phantom{$^{\ast\ast}$} & ~~\lose{46}\phantom{$^{\ast\ast}$} & ~~\lose{50}\phantom{$^{\ast\ast}$} \\
 & Arranger & ~~\lose{24}$^{\ast\ast}$ & ~~\win{54}\phantom{$^{\ast\ast}$} & ~~--\phantom{$^{\ast\ast}$} & ~~\lose{39}$^{\ast}$\phantom{$^{\ast}$} \\
 & Rewriter & ~~\lose{~9~}$^{\ast\ast}$ & ~~\lose{50}\phantom{$^{\ast\ast}$} & ~~\win{61}$^{\ast}$\phantom{$^{\ast}$} & ~~--\phantom{$^{\ast\ast}$} \\
\end{tabular}
\caption{Engagingness.}\label{fig:heengaging}
\end{subfigure}
\begin{subfigure}[b]{.46\linewidth}
\setlength{\tabcolsep}{3pt}
\centering
\scriptsize
\begin{tabular}{rr|cccc}
& & \multicolumn{4}{c}{Win \%}\\
& &  SimpleTOD & SimpleTOD+ & Arranger & Rewriter \\
\midrule
\parbox[t]{2mm}{\multirow{4}{*}{\rotatebox[origin=c]{90}{Loss \%}}}
& SimpleTOD & ~~--\phantom{$^{\ast\ast}$} & ~~\win{63}$^{\ast\ast}$ & ~~\win{73}$^{\ast\ast}$ & ~~\win{70}$^{\ast\ast}$ \\
 & SimpleTOD+ & ~~\lose{37}$^{\ast\ast}$ & ~~--\phantom{$^{\ast\ast}$} & ~~\lose{47}\phantom{$^{\ast\ast}$} & ~~\lose{50}\phantom{$^{\ast\ast}$} \\
 & Arranger & ~~\lose{27}$^{\ast\ast}$ & ~~\win{53}\phantom{$^{\ast\ast}$} & ~~--\phantom{$^{\ast\ast}$} & ~~\lose{40}$^{\ast}$\phantom{$^{\ast}$} \\
 & Rewriter & ~~\lose{30}$^{\ast\ast}$ & ~~\lose{50}\phantom{$^{\ast\ast}$} & ~~\win{60}$^{\ast}$\phantom{$^{\ast}$} & ~~--\phantom{$^{\ast\ast}$} \\
\end{tabular}
\caption{Interestingness.}\label{fig:heinteresting}
\end{subfigure}
\begin{subfigure}[b]{.46\linewidth}
\setlength{\tabcolsep}{3pt}
\centering
\scriptsize
\begin{tabular}{rr|cccc}
& & \multicolumn{4}{c}{Win \%}\\
& &  SimpleTOD & SimpleTOD+ & Arranger & Rewriter \\
\midrule
\parbox[t]{2mm}{\multirow{4}{*}{\rotatebox[origin=c]{90}{Loss \%}}}
& SimpleTOD & ~~--\phantom{$^{\ast\ast}$} & ~~\win{64}$^{\ast\ast}$ & ~~\win{77}$^{\ast\ast}$ & ~~\win{81}$^{\ast\ast}$ \\
 & SimpleTOD+ & ~~\lose{36}$^{\ast\ast}$ & ~~--\phantom{$^{\ast\ast}$} & ~~\lose{47}\phantom{$^{\ast\ast}$} & ~~\win{55}\phantom{$^{\ast\ast}$} \\
 & Arranger & ~~\lose{23}$^{\ast\ast}$ & ~~\win{53}\phantom{$^{\ast\ast}$} & ~~--\phantom{$^{\ast\ast}$} & ~~\lose{45}\phantom{$^{\ast\ast}$} \\
 & Rewriter & ~~\lose{19}$^{\ast\ast}$ & ~~\lose{45}\phantom{$^{\ast\ast}$} & ~~\win{55}\phantom{$^{\ast\ast}$} & ~~--\phantom{$^{\ast\ast}$} \\
\end{tabular}
\caption{Knowledge.}\label{fig:heknowledgable}
\end{subfigure}
\begin{subfigure}[b]{.46\linewidth}
\setlength{\tabcolsep}{3pt}
\centering
\scriptsize
\begin{tabular}{rr|cccc}
& & \multicolumn{4}{c}{Win \%}\\
& &  SimpleTOD & SimpleTOD+ & Arranger & Rewriter \\
\midrule
\parbox[t]{2mm}{\multirow{4}{*}{\rotatebox[origin=c]{90}{Loss \%}}}
& SimpleTOD & ~~--\phantom{$^{\ast\ast}$} & ~~\win{68}$^{\ast\ast}$ & ~~\win{71}$^{\ast\ast}$ & ~~\win{82}$^{\ast\ast}$ \\
 & SimpleTOD+ & ~~\lose{32}$^{\ast\ast}$ & ~~--\phantom{$^{\ast\ast}$} & ~~\win{51}\phantom{$^{\ast\ast}$} & ~~\lose{48}\phantom{$^{\ast\ast}$} \\
 & Arranger & ~~\lose{29}$^{\ast\ast}$ & ~~\lose{49}\phantom{$^{\ast\ast}$} & ~~--\phantom{$^{\ast\ast}$} & ~~\lose{40}$^{\ast}$\phantom{$^{\ast}$} \\
 & Rewriter & ~~\lose{18}$^{\ast\ast}$ & ~~\win{52}\phantom{$^{\ast\ast}$} & ~~\win{60}$^{\ast}$\phantom{$^{\ast}$} & ~~--\phantom{$^{\ast\ast}$} \\
\end{tabular}
\caption{Humanness.}\label{fig:hehumanlike}
\end{subfigure}
\caption{Human evaluation results on the test set of {\dn}-SGD using ACUTE-Eval ($^{\ast}$: $\text{p-value}<0.05$, $^{\ast\ast}$: $\text{p-value}<0.005$).} %
\label{fig:huamevals}
\end{figure*}

\begin{figure}[t!]
\setlength{\tabcolsep}{3pt}
\centering
\scalebox{0.76}{ 
\begin{tabular}{lccc}
 & {MultiWOZ 2.1} & \vs & {{\dn}-MultiWOZ} \\
 & (Win \%) & & (Win \%) \\
\midrule
 Engagingness & \lose{10}$^{\ast\ast}$ &  & \win{90}$^{\ast\ast}$ \\
 Interestingness & \lose{11}$^{\ast\ast}$ &  & \win{89}$^{\ast\ast}$ \\
 Knowledge & \lose{13}$^{\ast\ast}$ &  & \win{87}$^{\ast\ast}$ \\
 Humanness & \lose{10}$^{\ast\ast}$ &  & \win{90}$^{\ast\ast}$ \\
\end{tabular}
}
\caption{Comparisons between MultiWOZ 2.1 and {\dn}-MultiWOZ at the dataset level using ACUTE-Eval ($^{\ast\ast}$: $\text{p-value}<0.005$).} %
\label{fig:multiwoz}
\end{figure}

\paragraph{{\dn}-MultiWOZ.} To investigate the flexibility of our data construction approach, we augment about $1$K randomly sampled dialogues from another task-oriented dataset, MultiWOZ 2.1~\cite{eric2019multiwoz} following the same steps as described in Section~\ref{sec:data}. Crowd workers label $30.0\%$ of the candidates as good, which is lower compared with {\dn}-SGD ($41.4\%$ in Table~\ref{tab:statjustification}). We attribute the difference to (i) the performance downgrade of the filter model since we do not re-train the model for MultiWOZ 2.1, and (ii) a higher chance of a chit-chat augmented response being too verbose to be good since the average number of tokens per system turn in MultiWOZ 2.1 is larger than that of SGD ($17.3$ \vs $13.1$). Nevertheless, the augmented version (denoted as {\dn}-MultiWOZ) is significantly more preferred than the original, as shown in Figure~\ref{fig:multiwoz}, where we randomly sample $100$ dialogues from {\dn}-MultiWOZ, augment all of their system responses that have chit-chat candidates labeled as good, and compare these augmented dialogues with the corresponding original dialogues.

\subsection{Model Evaluations}
\label{sec:modeleval}

\paragraph{Automatic Evaluations.} We consider joint goal accuracy (Joint GA) and average goal accuracy (Avg GA) for evaluating belief states, act-slot F1 for evaluating action decisions, and two BLEU-4 scores (BLEU-4\textsubscript{orig}, BLEU-4\textsubscript{aug}) for evaluating system responses, where we use \textbf{orig}inal (\resp \textbf{aug}mented) system responses as references for BLEU-4\textsubscript{orig} (\resp BLEU-4\textsubscript{aug}). Table~\ref{tab:autoeval} summarizes the evaluation results. Since the test set of SGD contains unseen services (\ie, services not seen during training) designed to evaluate the model's generalizability, we report the results on all services (All) and seen services only (Seen) following \citet{rastogi2019towards}. Our proposed models generally achieve a similar task performance level compared with the SimpleTOD baseline. Unsurprisingly, the proposed models achieve lower BLEU-4\textsubscript{orig} and higher BLEU-4\textsubscript{aug}.

\paragraph{Human Evaluations.} We turn to human evaluations for a more comprehensive measure of the response generation performance. We employ the same ACUTE-Eval metrics as we do in data evaluations. We randomly sample $100$ dialogues from the test set of {\dn}-SGD. For each sampled dialogue $D=\{u_1,s_1,u_2,s_2,\ldots,u_n,s_n\}$, we pass $u_1,s_1,\ldots,u_i$ to each model $\mathcal{M} \in \{$SimpleTOD, SimpleTOD+, Arranger, Rewriter$\}$ to obtain its system response $s^{\mathcal{M}}_i$ for the $i$-th system turn ($1 \leq i\leq n$). Let $D^\mathcal{M}$ represent $\{u_1,s^{\mathcal{M}}_1,\ldots,u_n,s^{\mathcal{M}}_n\}$. We ask evaluators to compare each pair of $D^{\mathcal{M}_1}$ and $D^{\mathcal{M}_2}$, where $\mathcal{M}_1, \mathcal{M}_2\in \{$SimpleTOD, SimpleTOD+, Arranger, Rewriter$\}$ and $\mathcal{M}_1 \neq \mathcal{M}_2$. As shown in Figure~\ref{fig:huamevals}, all of the chit-chat augmented models outperform the SimpleTOD baseline over four ACUTE-Eval metrics. Among the chit-chat augmented models, no one shows a clear win over the other two on the quantitative level. We show a full dialogue example comparing responses generated by different models along with supplementary  discussions in Appendix~\ref{sec:samples}.

\begin{figure}[t!]
\setlength{\tabcolsep}{3pt}
\centering
\scalebox{0.76}{
\begin{tabular}{lccc}
 & SimpleTOD & \vs & Modified Arranger \\ %
 & (Win \%) & & (Win \%) \\
\midrule
 Engagingness & \lose{14}$^{\ast\ast}$ $\downarrow$10 &  & \win{86}$^{\ast\ast}$ $\uparrow$10 \\
 Interestingness & \lose{25}$^{\ast\ast}$ $\downarrow$2\phantom{0} &  & \win{75}$^{\ast\ast}$ $\uparrow$2\phantom{0} \\
 Knowledge & \lose{20}$^{\ast\ast}$ $\downarrow$3\phantom{0} &  & \win{80}$^{\ast\ast}$ $\uparrow$3\phantom{0} \\
 Humanness & \lose{20}$^{\ast\ast}$ $\downarrow$9\phantom{0} &  & \win{80}$^{\ast\ast}$ $\uparrow$9\phantom{0} \\
\end{tabular}
}
\caption{Human evaluation results of the modified Arranger with controlled injection frequency ($^{\ast\ast}$: $\text{p-value}<0.005$, $\uparrow$/$\downarrow$: increased/decreased win \% compared with the original Arranger).} %
\label{fig:injfreqctrl}
\end{figure}

\begin{table*}[ht!]
\centering
\scalebox{0.8}{
\begin{tabular}{lcccc}
\toprule
\bf Dataset & \bf Construction Method & \bf \# Dialogues & \bf Task-Oriented & \bf Chit-Chat \\
\midrule
DSTC2~\cite{henderson2014second} & crowdsourcing & $3,235$ & \cmark & \xmark \\
MultiWOZ 2.1~\cite{eric2019multiwoz} & crowdsourcing & $10,438$ & \cmark & \xmark \\
Schema-Guided Dialogue~\cite{rastogi2019towards}  & crowdsourcing & $22,825$ & \cmark & \xmark \\
SIMMC~\cite{moon2020situated}  & crowdsourcing & $12,948$ & \cmark & \xmark \\
\midrule
PersonaChat~\cite{zhang2018personalizing} & crowdsourcing & $10,907$ & \xmark & \cmark \\
Wizard of Wikipedia~\cite{dinan2018wizard} & crowdsourcing & $22,311$ & \xmark & \cmark \\
EmpatheticDialogues~\cite{rashkin2018towards} & crowdsourcing & $24,850$ & \xmark & \cmark \\
BlendedSkillTalk~\cite{smith2020can} & crowdsourcing & $6,808$ & \xmark & \cmark \\
Pushshift Reddit~\cite{baumgartner2020pushshift} & crawling \& scraping & $651,778,198^\dag$ & \xmark & \cmark \\
\midrule
\bf {\dn}-SGD (this work) & crowdsourcing & $22,825$ & \cmark  & \cmark \\
\bf {\dn}-MultiWOZ (this work) & crowdsourcing & $997$ & \cmark  & \cmark \\%-1K
\bottomrule
\end{tabular}
}
\caption{Statistics of dialogue datasets ($^\dag$: regarding each thread (\ie, a post and its comments) as a dialogue).}
\label{tab:related}
\end{table*}

Considering that the injection frequency affects human evaluations (Section~\ref{sec:dataeval}) and that all our models do not explicitly control the injection frequency, we experiment with controlling the injection frequency by modifying Arranger to consider including chit-chat into the current turn only when the injection frequency from the first turn to the current turn is less than $0.3$. Compared with the original Arranger, the modified Arranger achieves a higher win percentage over SimpleTOD, as shown in Figure~\ref{fig:injfreqctrl}. We leave further exploration of injection frequency for future work.

\subsection{Limitations and Further Discussions}

\paragraph{Approach.} Our proposed strategy to augment task-oriented dialogue system responses with chit-chat is simple, compared with how it emerges in human conversations, where both functionality and engagingness structurally intertwine with each other in a more complex fashion. Our proposed Rewriter model does have a modeling capability to compose both functions organically but is limited due to the dataset's target arrangement (\ie, concatenation of two separate components).  Despite the limitation, our chosen design of ``code-separation'' has practical merits: we can easily extend the proposed approach to an existing production-level virtual assistant system as a modularized solution, and it has minimal interference to the user-perceived \emph{task success rate}, a core metric widely adapted in virtual assistant systems. Another limitation of our work is that we only augment responses on the system side in our dataset, and the augmentations are independent of each other, whereas in real-life situations, users are also likely to make chit-chat, and the chit-chat between the user and the system should ideally be related to each other. We leave for future research addressing these limitations.

\paragraph{Evaluation.} We follow the previous literature on evaluation and regard the four ACUTE-Eval metrics as the primary measure of the response generation performance in this work. However, there is a large overlap between the desired quality measured by different human judgment categories used in ACUTE-Eval. The four ACUTE-Eval metrics favor the same dialogue $84.4\%$ of the time in our evaluation, indicating high correlations between these metrics. We leave the study of addressing this issue for future work.

\section{Related Work}
\label{sec:related}

\subsection{Dialogue Datasets}
\label{sec:relateddataset}
Dialogue system research has been consistently supported by the development of new datasets.  The Dialog State Tracking Challenge (DSTC) series~\cite{williams2013dialog,henderson2014second,henderson2014third,williams2014dialog,kim2016fifth,kim2017fourth,moon2020situated} provide common testbeds for task-oriented dialogues. Following DSTC, researchers have created a variety of publicly available task-oriented dialogue datasets~\cite{el-asri2017frames,shah2018building,budzianowski2018large,rastogi2019towards}. 
Another line of work seeks to facilitate open-domain chatbot development with large amounts of human-created text data generated in a social context~\cite{baumgartner2020pushshift} and supervision for a variety of desirable general qualities such as being engaging, personable, knowledgeable, and empathetic~\cite{zhang2018personalizing,dinan2018wizard,rashkin2018towards,moon2019opendialkg,wang2019persuasion,smith2020can}. Our work bridges the two lines. We compare {\dn}-SGD and {\dn}-MultiWOZ with relevant and representative dialogue datasets in Table~\ref{tab:related}.

Note that very few dialogue 
corpora contain explicit annotations for both task-oriented and chit-chat utterances. For example, task-oriented dialogue corpora constructed by \citet{rastogi2019towards} and \citet{moon2020situated} contain annotations for a few chit-chat dialogue acts, but they are limited to light social greetings (\eg, ``\textit{Thank you!}'', ``\textit{Good Bye.}'') %
typically at the end of each dialogue session.
\citet{zhao2017generative} propose to artificially augment task-oriented dialogues with randomly sampled utterances from a chit-chat corpus, mainly to improve the out-of-domain recovery performance. \citet{akasaki2017chat} annotate user utterances with chat/non-chat binary labels. Still, they do not study the contextual combination of these two to make conversations more engaging, and their corpus does not contain goal labels like typical task-oriented dialogue corpora. In contrast, our work drastically increases the diversity and contextual coverage of chit-chat additions for any task-oriented dialogue corpus (\eg, ``\textit{It's a great way to kick off the summer!}'', ``\textit{I hear it's beautiful.}'').

Compared with other approaches of creating a high-quality dialogue corpus (\eg, via human-to-human ``Wizard-of-Oz'' collection~\cite{eric2019multiwoz}, dialogue self-play and paraphrase~\cite{shah2018building}), the annotation cost of the proposed model-based dialogue generation approach combined with the quality control mechanisms is lower, as our work does not involve authoring new sentences by human annotators. 

\subsection{Task-Oriented Dialogue Systems}

Over the past few years, neural models have achieved remarkable success in the development of the main components of task-oriented dialogue systems, including understanding user intent, tracking dialogue states, determining system actions, and generating system responses~\cite{henderson2013deep,sun2014sjtu,wen2015semantically,liu2016attention,mrksic2017neural,wen2017latent,nouri2018gce,heck2020trippy,chen2020schema}. Recently, connecting separate components and building end-to-end task-oriented neural dialogue systems have attracted increasing interest~\cite{bordes2016learning,peng2020few}. The most recent thread is to unify all components in a single end-to-end neural model by fine-tuning a pre-trained deep language model on multiple tasks%
, which leads to state-of-the-art performance~\cite{hosseini2020simple,peng2020soloist}. We follow this thread and further enhance the ability to generate appropriate non-task-oriented add-ons, on top of the ability to achieve functional goals that existing systems are typically narrowly tailored to. 
A few work have studied training a dialogue model leveraging multiple chit-chat and task-oriented dialogues~\cite{madotto2020attention,madotto2020adapter}, which allows the model to attend on a relevant task for a given user utterance and respond accordingly, thus increasing the skill coverage of the model. 
Our proposed models are trained on the newly collected {\dn}-SGD dataset with the turn-level supervision signals, allowing for contextual and flexible code-switching between chit-chat and functional tasks in a single system turn.

\section{Conclusion}

We propose adding chit-chat to enhance task-oriented dialogues ({\dn}) in this study. We present a general Human$\leftrightarrow$AI collaborative data construction approach for {\dn}, with which we create a dataset consisting of $23.8$K chit-chat augmented task-oriented dialogues. We show via human evaluation that chit-chat augmented dialogues are preferred than the unaugmented. In addition, we propose three models for {\dn}. Evaluation results show that compared with the baseline trained on the original unaugmented data, our proposed models trained on the chit-chat augmented counterpart achieve a similar task performance level and higher human evaluation scores.

\section*{Acknowledgements}
We thank Gerald Demeunynck for helping with the data annotation process.
We would also like to thank the anonymous NAACL reviewers for their constructive and insightful feedback.
\pagebreak

\bibliography{naacl2021}
\bibliographystyle{acl_natbib}

\appendix

\clearpage

\section{Appendix}
\label{sec:appendix}

\subsection{Details of Candidate Generation}
\label{sec:candsgendetails}

We summarize model configurations in Table~\ref{tab:modelconfigs}, which are employed together for candidate generation in Section~\ref{sec:candgen}. Our implementation is based on ParlAI~\cite{miller2017parlai}, and all unspecified parameters take the default values set in the interactive mode of ParlAI.

\begin{table}[ht!]
\centering
\scriptsize
\begin{tabular}{lcc}
\toprule
 \bf Generative Model & \bf Beam Size & \bf Minimum Beam Length \\ 
 \midrule
BlenderBot ($90$M) & $10$ & $1$ \\
BlenderBot ($90$M) & $10$ & $5$ \\
BlenderBot ($90$M) & $10$ & $20$ \\
BlenderBot ($90$M) & $30$ & $20$ \\
BlenderBot ($2.7$B) & $10$ & $1$ \\
BlenderBot ($2.7$B) & $10$ & $5$ \\
BlenderBot ($2.7$B) & $10$ & $20$ \\
BlenderBot ($2.7$B) & $30$ & $20$ \\
BlenderBot ($9.4$B) & $10$ & $1$ \\
BlenderBot ($9.4$B) & $10$ & $20$ \\
GPT-2 ($117$M) & $10$ & $1$ \\
GPT-2 ($345$M) & $1$ & $1$ \\
GPT-2 ($345$M) & $10$ & $1$ \\
GPT-2 ($345$M) & $10$ & $5$ \\
GPT-2 ($345$M) & $10$ & $20$ \\
GPT-2 ($345$M) & $30$ & $1$ \\
GPT-2 ($762$M) & $10$ & $1$ \\
\bottomrule
\end{tabular}
\caption{Employed models and decoding parameters for candidate generation.}
\label{tab:modelconfigs}
\end{table}

\subsection{Details of Candidate Filtering}
\label{sec:filterdetails}
The ranker initially ranks each candidate according to the posterior probability output by the binary classifier. It then lowers the ranks of candidates that match a list of bad patterns. Most bad patterns are about newly introduced counterfeit information (\eg, containing an URL/email address, a phone number, time, or amount of money). The rest of the bad patterns are mainly about text genre (\eg, containing email sign-offs such as ``best regards'') and format (\eg, misuse of punctuation marks). Lastly, the ranker raises the ranks of (i) uncommon candidates and (ii) candidates that are dissimilar to the other candidates for the dialogue and the system response being augmented. We measure the similarity by Levenshtein distance.

Note that we do not explore the optimal settings for candidate filtering, as it is not the primary focus of this paper. For instance, how much the rule-based ranker lowers or raises the ranks of candidates is set manually based on engineering intuition rather than rigorous analysis; we do not exhaustively investigate how much labeled data is required to obtain a good enough binary classifier; the $1.7$K examples from the pilot annotation are randomly sampled. Tuning the procedure (\eg, the number and selection of training examples) may lead to a better resulting candidate set.

\subsection{Human Evaluation Questions}
\label{sec:acuteevalqs}

\begin{itemize}
    \item Engaging: \textit{Who would you prefer to talk to? Which version is more likely to hold your attention and make you want to hear more?}
    \item Interesting: \textit{Who would you say is more interesting? Which version arouses your curiosity or tells you something new or useful?}
    \item Humanlike: \textit{Who would you say sounds more human? Which version is more natural and personable?}
    \item Knowledgeable: \textit{Who would you say is more knowledgeable? Which version seems more well informed and confident in the information?}
\end{itemize}

\clearpage

\subsection{Example Dialogues}
\label{sec:dataset_examples}

\begin{minipage}{\textwidth}
\centering
\scriptsize
\captionsetup{type=table}
\begin{tabular}{p{0.2cm}p{8.7cm}p{5.8cm}}
\toprule
 & \textbf{Utterance} & \textbf{Task Labels} \\
\midrule
\textbf{U}: & I'm starting to get hungry. I want to go eat at a nice restaurant.  & INFORM\_INTENT [intent: FindRestaurants]  \\
\textbf{A}: & Sure. What city would you like to eat in?  & REQUEST [slot: city]  \\
\textbf{U}: & Milpitas.  & INFORM [city: Milpitas] \\
\textbf{A}: & (\xmark) \textit{\ul{That sounds delicious.}}  &  \\
            & Do you have a particular type of food you are craving? Mexican? American? Something else?  & REQUEST [cuisine: Mexican, American] \\
\textbf{U}: & I feel like Cambodian tonight.  & INFORM [cuisine: Cambodian]  \\
\textbf{A}: & (\cmark) \textit{\ul{I hear the food is amazing.}}  & OFFER [restaurant\_name: Anh Hong, city: Milpitas] \\
            & There is a Cambodian restaurant in Milpitas called Anh Hong.  &  \\
            & (\cmark) \textit{\ul{It's a great place to eat.}}  &  \\            
\textbf{U}: & That's exactly what I'm looking for. & SELECT \\
\textbf{A}: & Do you want me to make a reservation for you?  & OFFER\_INTENT [intent: ReserveRestaurant] \\
            & (\xmark) \textit{\ul{I can also recommend some other restaurants in the area.}}  &  \\            
\textbf{U}: & Yes, I'd like to try this place out. & AFFIRM\_INTENT \\
\textbf{A}: & Okay. At what time would you like to eat there? & REQUEST [slot: time]  \\
\textbf{U}: & 6:30 pm. Please make the reservation for four people. & INFORM [time: 6:30 pm, party\_size: 4] \\
\textbf{A}: & (\xmark)  \textit{\ul{Okay}} . & CONFIRM [restaurant\_name: Ahn Hong,  \\
            & Sure. A reservation at Anh Hong in Milpitas for 4 at 6:30 pm today. Is that correct? & city: Milpitas, time: 6:30 pm, party\_size: 4]  \\        
\textbf{U}: & Perfect. What are the prices at Anh Hong like? &  REQUEST [slot: price\_range]\\
\textbf{A}: & They have moderate pricing. Your reservation has been made. & INFORM [price\_range: moderate] \\
            & (\cmark)  \textit{\ul{Thank you.}} &  \\        
            & (\xmark)  \textit{\ul{Do you want dessert?}} &  \\           
\textbf{U}: & Thanks. You've been a big help. & THANK\_YOU \\
\textbf{A}: & (\cmark)  \textit{\ul{You are welcome. I hope you enjoy your meal.}} & GOODBYE \\
            & Have a nice day. \\           
            & (\xmark)  \textit{\ul{I hope you enjoy your meal.  Do you have any plans for the rest of the day?}} \\           
\midrule
\textbf{U}: & Can you find me some songs? Any album is fine. & INFORM\_INTENT [intent: LookupMusic, album: dontcare] \\
\textbf{A}: & I have found 10 songs you may like. How does A Little Bit Stronger by Sara Evans in the album Stronger sound? & INFORM\_COUNT [count: 10], OFFER [artist: Sara Evans, song\_name: A Little Bit Stronger] \\
\textbf{U}: & That's great. & SELECT  \\
\textbf{A}: & (\xmark) \textit{\ul{I'll have to check it out.}} & REQ\_MORE  \\
            & (\cmark) \textit{\ul{I'm glad you like it.}} & \\         
            & What else can I help with? & \\           
            & (\cmark) \textit{\ul{I love helping people find new music.}} & \\         
\textbf{U}: & I want to play the song. &  INFORM\_INTENT [intent: PlayMedia] \\
\textbf{A}: & Should I play A Little Bit Stronger on the TV? & CONFIRM [playback\_device: TV,  \\
            & (\cmark) \textit{\ul{I think it would be a great song to play on the television. What do you think?}} & song\_name: A Little Bit Stronger] \\ 
            & (\cmark) \textit{\ul{That's my favorite song by Sarah Evans.}} & \\             
\textbf{U}: & Yes, that's correct. & AFFIRM \\
\textbf{A}: & Started playing the song. & NOTIFY\_SUCCESS \\
            & (\xmark) \textit{\ul{I can tell you about the lyrics.}} & \\ 
            & (\xmark) \textit{\ul{What else do you want to know about it?}} & \\             
\textbf{U}: & Thanks, that's all I need. & THANK\_YOU \\
\textbf{A}: & (\cmark) \textit{\ul{You're welcome.}} & GOODBYE \\
            & Have a good day. & \\ 
            & (\xmark) \textit{\ul{What do you like?}} & \\ 
            & (\xmark) \textit{\ul{I hope you find some good music to listen to. I like all kinds of music. What do you like?}} & \\             
\bottomrule
\end{tabular}
\captionof{table}{Example dialogues from \dn-SGD (U: user; A: assistant). All \textit{\ul{chit-chat}} candidates for augmentation, generated with the state-of-the-art pre-trained language models, are annotated by the crowd workers with good (\cmark) and (\xmark) bad labels. Note that while most of the bad chit-hat candidates are \textit{fluent}, they are often contextually inappropriate or inconsistent with the rest of the dialogue. The annotation guideline is highlighted in Section~\ref{sec:annotation}.} 
\label{tab:dataset_examples}
\end{minipage}

\clearpage

\subsection{Generation Samples}
\label{sec:samples}
\begin{minipage}{\textwidth}
\centering
\scriptsize
\captionsetup{type=table}
\begin{tabular}{p{0.2cm}p{2.3cm}p{0.2cm}p{11.7cm}}
\toprule
\textbf{U}: & I'm looking for something interesting to do, please find an activity for me on the 5th of this month. & \textbf{ST}: \newline\newline \textbf{S+}: \newline\newline \textbf{AR}: \newline\newline \textbf{RW}: & What type of events do you prefer, maybe [events\_category] or [events\_category] or anything else? Also, what city should I search in? \newline Sure, I can assist you. So, where are you looking for an activity, and do you have a type of activity you prefer? You could say music, sports, or something like that. \newline What type of events do you prefer, maybe [events\_category] or [events\_category] or anything else? Also, what city should I search in? \newline What type of events do you prefer, maybe [events\_category] or [events\_category] or anything else? Also, what city should I search in? \\
\hdashline
\textbf{U}: & I'd like to search for an activity around Seattle, WA. & \textbf{ST}:\newline \textbf{S+}: \newline \textbf{AR}: \newline \textbf{RW}: & Is there any particular activity you are interested in? \newline What type of activity would you like to take part in? There are a lot of activities to be found in the city. \newline Seattle is a great place to visit. Is there any particular activity you are interested in? \newline Seattle is a great place to visit. Is there any particular activity you are interested in? \\
\hdashline
\textbf{U}: & I'd like to do something Music related. & \textbf{ST}: \newline\newline \textbf{S+}:\newline\newline \textbf{AR}: \newline\newline \textbf{RW}: & I found [events\_count] events that you might like. There's a [events\_event\_name] activity taking place at [events\_event\_location]. It's happening [events\_date] starting at [events\_time]. \newline I discovered [events\_count] activities you may enjoy. There is an [events\_event\_name] activity at [events\_event\_location] on [events\_date], starting at [events\_time]. \newline I found [events\_count] events that you might like. There's a [events\_event\_name] activity taking place at [events\_event\_location]. It's happening [events\_date] starting at [events\_time]. \newline It's a great place to see a concert. I found [events\_count] events that you might like. There's a [events\_event\_name] activity taking place at [events\_event\_location]. It's happening [events\_date] starting at [events\_time].\\
\hdashline
\textbf{U}: & That seems fun, please book one ticket for me to that event. & \textbf{ST}: \newline\newline \textbf{S+}: \newline\newline \textbf{AR}: \newline\newline \textbf{RW}: & Confirm the following details for me: Booking [events\_number\_of\_seats] ticket for the [events\_event\_name] show on [events\_date] in [events\_city\_of\_event]. \newline Let me see if I got everything correctly: You'd like [events\_number\_of\_seats] ticket for the [events\_event\_name] show in [events\_city\_of\_event] on [events\_date]. \newline Confirm the following details for me: Booking [events\_number\_of\_seats] ticket for the [events\_event\_name] show on [events\_date] in [events\_city\_of\_event]. \newline I'm sure you'll enjoy it. Confirm the following details for me: Booking [events\_number\_of\_seats] ticket for the [events\_event\_name] show on [events\_date] in [events\_city\_of\_event].\\
\hdashline
\textbf{U}: & Yes, that is correct. Please tell me the venue's address. & \textbf{ST}: \newline \textbf{S+}: \newline \textbf{AR}: \newline \textbf{RW}: & I have reserved your ticket. The venue is located at [events\_address\_of\_location]. \newline Reservation is confirmed. The venue is located at [events\_address\_of\_location]. \newline I have reserved your ticket. The venue is located at [events\_address\_of\_location]. \newline I have reserved your ticket. The venue is located at [events\_address\_of\_location].\\
\hdashline
\textbf{U}: & Okay, please find me a ride to the event. A shared ride is fine. & \textbf{ST}: \newline\newline \textbf{S+}: \newline \textbf{AR}: \newline\newline \textbf{RW}: & Please confirm that you would like to travel to [ridesharing\_destination] in a cab for [ridesharing\_number\_of\_riders] person and that you would like a shared ride. \newline How many people will be riding in the cab? \newline Please confirm that you would like to travel to [ridesharing\_destination] in a cab for [ridesharing\_number\_of\_riders] person and that you would like a shared ride. \newline  Please confirm that you would like to travel to [ridesharing\_destination] in a cab for [ridesharing\_number\_of\_riders] person and that you would like a shared ride.\\
\hdashline
\textbf{U}: & Yes that is correct. How long will it take to get there. & \textbf{ST}:\newline \textbf{S+}: \newline \textbf{AR}: \newline \textbf{RW}: & Your ride is booked and on the way. The ride is expected to take [ridesharing\_approximate\_ride\_duration] minutes. \newline Your ride is confirmed and is on its way. The ride is expected to take [ridesharing\_approximate\_ride\_duration] minutes. \newline Your ride is booked and on the way. The ride is expected to take [ridesharing\_approximate\_ride\_duration] minutes. \newline Your ride is booked and on the way. The ride is expected to take [ridesharing\_approximate\_ride\_duration] minutes.\\
\hdashline
\textbf{U}: & What's the cost of the ride? & \textbf{ST}: \newline \textbf{S}+: \newline \textbf{AR}: \newline \textbf{RW}: & The cost is [ridesharing\_ride\_fare]. \newline The price is [ridesharing\_ride\_fare]. \newline The cost is [ridesharing\_ride\_fare]. \newline The cost is [ridesharing\_ride\_fare]. \\
\hdashline
\textbf{U}: & Thanks, I don't need anything else. & \textbf{ST}:\newline \textbf{S+}: \newline \textbf{AR}: \newline \textbf{RW}: & Enjoy your day. \newline Enjoy the rest of your day. If you have any questions, please don't hesitate to contact me. \newline You are welcome. Enjoy your day. \newline You are welcome. Enjoy your day.\\
\bottomrule
\end{tabular}
\captionof{table}{Sampled responses generated for a complete task-oriented dialogue (U: user, ST: SimpleTOD, S+: SimpleTOD+, AR: Arranger, RW: Rewriter).} %
\vspace{10pt}
\label{tab:quadwise}
\end{minipage}
As shown in Table~\ref{tab:quadwise}, we observe that compared with SimpleTOD+, both Arranger and Rewriter tend to add chit-chat to the beginning of task-oriented responses. This is perhaps because the underlying off-the-shelf chit-chat model takes only $u_1,s_1,\ldots,u_i$ as input, making it more likely to generate a suitable chit-chat to start, rather than end the $i$-th system turn. The responses generated by Arranger and Rewriter are similar because Rewriter generates responses by copying contents from the responses output by the underlying off-the-shelf models without modification for most of the time ($87.0\%$ of dialogues on the test set).

\iffalse

\subsection{}

\begin{figure}[h!]
   \begin{center}
   \includegraphics[width=0.45\textwidth]{figure/seenunseen.pdf}
   \end{center}
%
 \caption{Human evaluation results on dialogues with seen, unseen, or mixed services using ACUTE-Eval.} %
 %
 \label{fig:seenunseen}
\end{figure}

To investigate the effect of the presence of the service in training data, we further perform human evaluation separately on dialogues that involve only seen services (Seen), only unseen services (Unseen), and both seen and unseen services (Mixed). We randomly sample $100$ dialogues in each of the three categories from the test set of {\dn}-SGD and compare SimpleTOD+ and SimpleTOD on these dialogues. Perhaps surprisingly \shane{I think we should either try to come up with a more clear explanation / hypothesis for this result -- or if we can't, it might be better to take out Figure 7 all together to make the main results more clear. My sense is that it is related to the automatic task performance of seen, unseen and mixed services -- iff that's the case we can move Figure 7 *and* add the automatic evaluation results breakdown to supplementary, and explain their correlations there.}, we notice that the relative improvement achieved by adding chit-chat is significant for Seen and Unseen ($\text{p-values}<0.005$) but not for Mixed, as shown in Figure~\ref{fig:seenunseen}. 

\fi

%
%
%

%
%

%
%
%
%
%
%
%
%
%
%

%

\end{document}